\documentclass{article}
\usepackage{graphicx}
\usepackage[T1]{fontenc}
\usepackage[utf8]{inputenc}
\usepackage{fvextra}
\usepackage{amsmath}
 \usepackage[preprint]{neurips_2026}

\usepackage[utf8]{inputenc} 
\usepackage[T1]{fontenc}    
\usepackage{hyperref}       
\usepackage{url}            
\usepackage{booktabs}       
\usepackage{amsfonts}       
\usepackage{nicefrac}       
\usepackage{microtype}      
\usepackage{xcolor}         

\title{Sparks of In Silico Cognitive Science: Theories from Simulated Data Can Generalize to Humans}

\workshoptitle{Sim2Science}

\author{
  Akshay K. Jagadish  $\dagger$  \\
  Princeton University \\
  \texttt{akshay.jagadish@princeton.edu} \\
  \And
  Younes Strittmatter  $\dagger$  \\
  Princeton University \\
  \And
  Nori Jacoby \\
  Cornell University \\
  \And
  Eric Schulz \\
  Helmholtz Munich \\
  \And
  Nathaniel Daw \\
  Princeton University \\
  \And
  Thomas L. Griffiths \\
  Princeton University \\
  \And
  Suyog H. Chandramouli \\
  Princeton University\\
  \And
  \\
  $\dagger$ Equal contribution
}

\begin{document}
\setlength{\textfloatsep}{6pt plus 2pt minus 2pt}

\maketitle

\begin{abstract}
Behavioral foundation models have been proposed as stand-ins for human participants across settings, but it is unclear whether theories discovered on them generalize to humans or merely characterize the simulator. We ran the Automated Cognitive Scientist (\textsc{AutoCog}), a closed-loop discovery system in which LLM agents design theory-discriminating experiments, collect responses, arbitrate between competing theories, and synthesize successors, entirely on behavior simulated by Centaur, a foundation model of human behavior. In a multi-attribute decision-making setting, the theories \textsc{AutoCog} found on Centaur generalized to human data: they outperformed canonical theories on ten held-out experiments and were rivaled only by theories found by running the same loop on people. We argue that this succeeds despite the simulator's inevitable imperfections because a discovery loop that arbitrates between competing theories demands less of its simulator than estimation does. The simulator only needs to capture the regularities that distinguish the theories, and not necessarily reproduce behavior precisely. Imperfect simulators can therefore widen the search over theories, with human data then testing whether the surfaced theories generalize.
\end{abstract}

Autonomous systems now contribute to scientific discovery across many disciplines \citep{abolhasaniRiseSelfdrivingLabs2023, novikov2025alphaevolve}. This includes psychology \citep{autocog2026, prystawski2026autopsych}, where LLM agents propose theories, design experiments that discriminate among them, collect behavior from online participants, and revise their theories to account for the empirical data. As these systems scale, recruitment budgets and participant time become key bottlenecks. Foundation models pretrained on a field's accumulated data offer a way around them, serving as surrogates when direct measurement is costly \citep{lin2023esm, bodnar2025aurora}, and a model trained on a large enough corpus can even approximate the underlying regularities better than direct empirical estimates do \citep{agrawal2020scaling}.
Centaur extends surrogate modeling to psychology by fine-tuning a language model on trial-level data from 160 experiments to predict human responses \citep{binzFoundationModelPredict2025}. A recent proposal goes further, envisioning an in silico science of the mind, in which the entire discovery cycle runs on such models and human studies are reserved for confirmation \citep{jagadishCanWeAutomatize2026}. More conservatively, a hybrid workflow could run the early cycles of theory search on synthetic responses and hand the later cycles to human participants. Whether theories discovered in these ways generalize to people or merely characterize the simulator \citep{vandergrift2026position} is an open question; we provide a direct test by running the \textsc{AutoCog} theory-discovery loop on Centaur alone.

\begin{figure}[t]
    \centering
    \includegraphics[width=.9\linewidth]{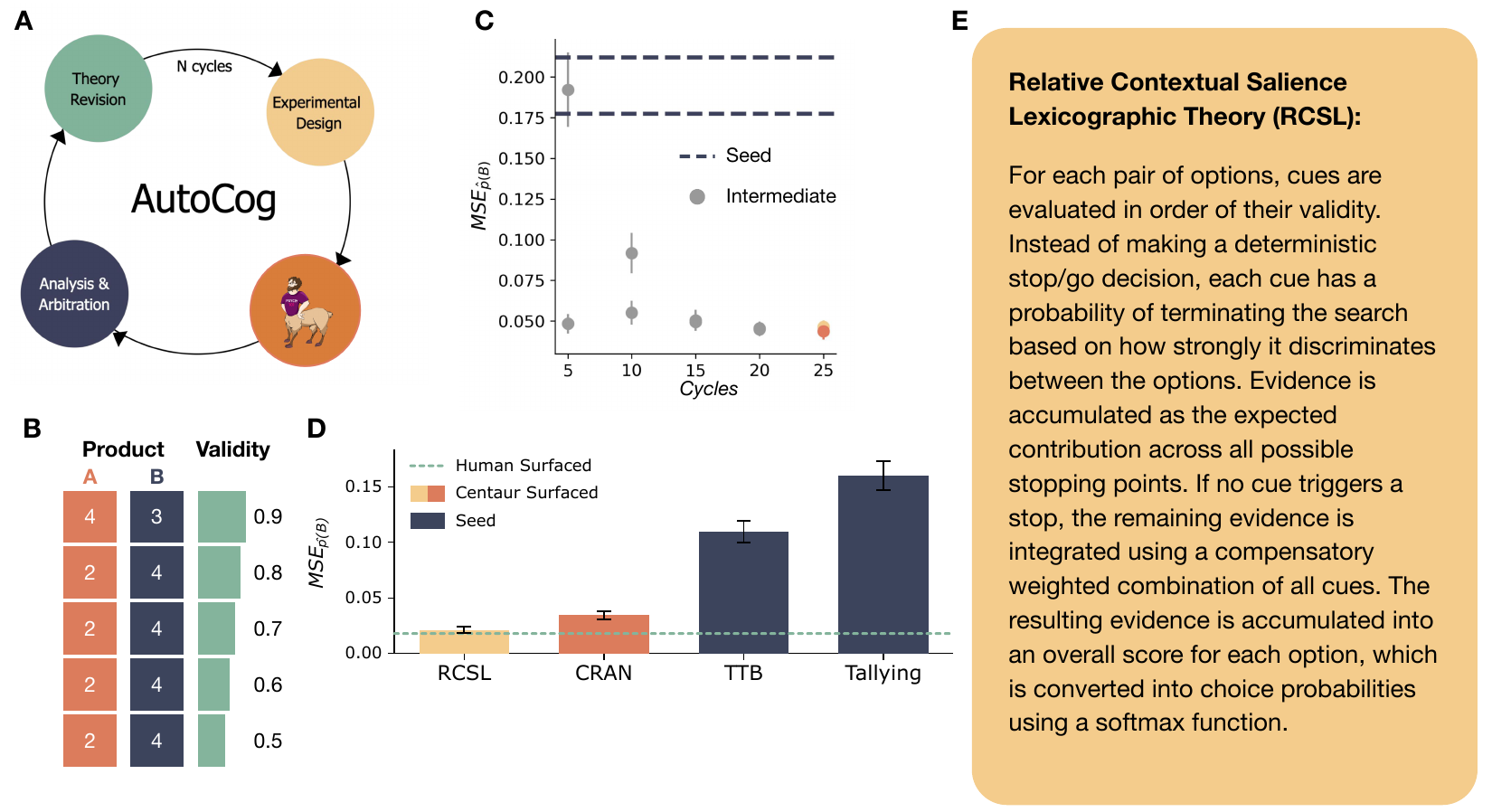}
    \caption{\textbf{A.} The \textsc{AutoCog} loop, with Centaur simulating human behavior. \textbf{B.} Task: choose between two products described by up to five cardinal-valued cues with validities 0.5--0.9. \textbf{C.} Mean squared error between a theory's and Centaur's proportions of B choices per option pair ($\mathrm{MSE}_{\hat{p}(B)}$) across cycles, with the two incumbent theories (gray), the two surfaced final theories (peach, gold) and the seeds (dashed).
    \textbf{D.} The same error against human choices on ten held-out experiments; the teal line marks the theories surfaced with human participants.
    \textbf{E.} The RCSL theory, which generalized best. Error bars: SEM over the fifty Centaur experiments (\textbf{C}) and over the ten human experiments (\textbf{D}).}
    \label{fig:msa}
    \vspace{-0.5mm}
\end{figure}

\textsc{AutoCog} \citep{autocog2026} runs a four-stage loop (Figure~\ref{fig:msa}A), starting from two seed theories; here we ran a single loop of 25 cycles. In \emph{experimental design},  an LLM agent designs one experiment per incumbent theory to discriminate it from its rival. In \emph{behavioral data collection}, Centaur responds to each experiment trial by trial in place of a human participant. In \emph{analysis and arbitration}, both theories simulate responses for each experiment and are scored by the divergence between their predictions and Centaur's responses, and an LLM-arbiter declares which theory the data favor. In \emph{theory revision}, the weaker theory is rewritten through program synthesis \citep{rmusGeneratingComputationalCognitive2025}, either by re-deriving its model or by proposing a new mechanism, so the loop searches an open space of mechanisms rather than a predefined set.


During the run, prediction error fell over cycles as successive theories replaced the seeds. The two final theories were a relative contextual salience lexicographic (RCSL) theory and a contextual relative advantage normalization (CRAN) theory. Both recombined familiar components from the literature into accounts absent from the seed set; RCSL, for instance, combines Take-The-Best's validity-ordered search with compensatory integration, where relative cue differences influence stopping and a validity-weighted sum determines the choice when no cue stops the search (Figure~\ref{fig:msa}E). Both reduced error on Centaur's behavior ($\mathrm{MSE}_{\hat{p}(B)}$, Figure~\ref{fig:msa}C) well below either seed. On ten held-out human experiments collected by \citet{autocog2026}, the surfaced theories (RCSL: $\mathrm{MSE}_{\hat{p}(B)} = 0.021 \pm 0.002$; CRAN: $0.034 \pm 0.004$; Figure~\ref{fig:msa}D) outperformed Take-The-Best ($0.110 \pm 0.010$) and Tallying ($0.161 \pm 0.013$) and were rivaled only by the theories \textsc{AutoCog} discovered with humans in the loop (Diminishing Returns WADD: $0.018 \pm 0.005$; Satisficing WADD: $0.018 \pm 0.003$; teal line in Figure~\ref{fig:msa}D), which were derived from these ten experiments.

Behavioral foundation models diverge from human behavior in places \citep{namazova2025not,xie2025centaur}, yet the theories discovered on Centaur predicted held-out human behavior better than the canonical seeds. These findings are compatible because a discovery loop demands less of its simulator than estimation \citep{peherstorfer2018survey}. Because the \textsc{AutoCog} agents restrict the search to plausible theories, simulated behavior enters the loop only through two comparisons: between incumbent theories during arbitration and between the weaker theory and the data during revision, so the simulator only needs to be faithful to human behavior along the dimensions these comparisons use. Centaur's errors could, in principle, shape successor theories during revision, yielding theories of Centaur rather than people—a failure mode that \textsc{AutoCog}'s arbitration cannot detect. Generalization to held-out human data indicates that this did not occur here, plausibly because the designed experiments stayed close enough to Centaur's training distribution (Appendix). Fidelity should matter more as theories converge in their predictions or as designed experiments depart from the simulator's training distribution \citep{liu2026centaur}.
A promising hybrid workflow might run early cycles on simulated data while candidate theories make clearly different predictions and experiments stay near the simulator's training distribution, but move to human participants once either condition fails. We take these results as sparks of an in silico cognitive science, where simulators expand the space of candidate theories and human data adjudicate among them.

\bibliographystyle{plainnat}
\bibliography{references}
\appendix

\section*{Appendix}

\subsection*{Distance from Centaur's training distribution}
Centaur's training corpus includes multi-attribute decision-making with binary cues, but not the cardinal-valued variant that we studied here; our experiments therefore lie near, though not within, its training distribution. Three features of the task plausibly explain why its simulations remained useful for discovery nonetheless. First, the task is a direct extension of the paradigms on which Centaur was fine-tuned -- specifically, the multi-attribute decision making task with binary valued features developed by \citet{hilbig2014generalized}. Second, choices receive no feedback and no correct answer is ever revealed, so there is no accuracy shortcut for the model to exploit. Third, trials are independent, so sequential dependencies, on which the model has been reported to diverge from people \citep{namazova2025not}, never arise. We nonetheless expect distance from the training distribution to matter in degree, and we identify it as one of the conditions under which we recommend turning to human participants.

\subsection*{Theory representation}
Every theory in \textsc{AutoCog}, seed or surfaced, pairs a verbal statement of a mechanism with an executable model that satisfies a fixed interface. This includes a \emph{Parameters} block declaring each free parameter and its admissible range (cue validities are passed in from the experiment), a \texttt{predict(parameters, stimulus, history)} function (called \texttt{generate} in \citealp{autocog2026}) returning choice probabilities over the two options, and a \texttt{policy} function mapping those probabilities to a response. All theories share a softmax choice rule with inverse temperature \texttt{beta} and a lapse rate \texttt{epsilon}, which is why these two parameters appear in every block. In the framework the interface is specified manually, and \textsc{AutoCog} produces the verbal description of the theory, the body of \texttt{predict}, and the parameter specification as well. When a theory is simulated, each synthetic participant draws its parameters once from the declared ranges, so theories are compared based on the behavior they generate (generative evaluation) rather than on a maximum-likelihood fit.

\subsection*{Run details}
All agents were driven by \texttt{gemini-3.1-pro-preview} through the \texttt{google-genai} SDK at sampling temperature 0.7, with a maximum output of 32{,}768 tokens and a reasoning budget of 8{,}096 tokens, as in \citet{autocog2026}. Centaur (Llama-3.1-Centaur-70B, run as the released adapter on a 4-bit Llama-3.1-70B base) supplied 25 synthetic participants per designed experiment, each completing about 96 trials with every unique option pair repeated $\max(1, \lfloor 96 / n_{\text{pairs}} \rfloor)$ times, in one random order shared across participants with the option labels randomized per participant; each choice was a single next token sampled at temperature 1.0 from a prompt in the text format of Centaur's training data. Theories were evaluated by replaying them on each experiment's option pairs, for the fifty Centaur experiments (Figure~\ref{fig:msa}C) and the ten held-out human experiments (Figure~\ref{fig:msa}D), across 200 simulated participants with parameters drawn once from their declared ranges (random seed 0), giving $\hat{p}(B)$ per pair; the same pipeline produced the held-out numbers for the human-loop theories, so those comparisons are like for like. Centaur inference ran on a single NVIDIA H200 GPU (141 GB) on an academic cluster; the full run took roughly 24 GPU-hours (estimated, not logged).

\subsection*{Seed heuristics}

\paragraph{Take-The-Best (TTB)}
\textit{Theory.} People compare two options by consulting cues one at a time in order of validity, stopping at the first cue that discriminates between the two options. That cue alone determines the choice: the option with the higher value on the discriminating cue wins, and no other cue is consulted. Cues with lower validity are never reached when a higher-validity cue already discriminates, so Take-The-Best is a ``one-reason'' decision rule --- only a single feature is ever used on any given choice. Because only the sign of the comparison on the top discriminating cue matters, TTB ignores both the magnitude of that difference and all information on lower-validity cues, making it maximally frugal in the use of evidence. When no cue discriminates (all feature-wise comparisons tie) the learner has no basis for preference and must guess. Response noise enters through a softmax over the binary TTB score (winner = 1, loser = 0) with inverse temperature beta, plus an independent lapse that with probability epsilon replaces the softmax output with a uniform pick over the two options.

\textit{Parameters.}
\begin{Verbatim}[breaklines=true,breakanywhere=true,fontsize=\footnotesize]
beta: [0.1, 20.0]
epsilon: [0.0, 0.5]
validities: validities
\end{Verbatim}

\textit{predict.}
\begin{Verbatim}[breaklines=true,breakanywhere=true,fontsize=\footnotesize]
def predict(parameters, stimulus, history):
    # Paper-faithful Take The Best (Gigerenzer & Goldstein 1996).
    # Stimulus is the pair of option feature vectors for the current
    # trial: array-like of shape (2, n_features), row 0 = option A,
    # row 1 = option B. Cue cascade: features are consulted in order
    # of descending validity; the first discriminating cue (strict
    # inequality) determines the winner; if no cue discriminates,
    # the model guesses uniformly. History is ignored.
    stim = np.asarray(stimulus, dtype=float)
    if stim.ndim != 2 or stim.shape[0] != 2:
        raise ValueError(
            f"TTB expects a (2, n_features) stimulus; got shape {stim.shape}."
        )
    n_features = stim.shape[1]

    val = np.asarray(parameters["validities"], dtype=float)
    if val.shape[0] != n_features:
        raise ValueError(
            f"validities length {val.shape[0]} != n_features {n_features}."
        )
    # Descending validity; argsort is stable so validity ties break
    # toward the earlier feature index.
    cue_order = np.argsort(-val, kind="stable").tolist()

    a, b = stim[0], stim[1]
    winner = None
    for j in cue_order:
        if a[j] > b[j]:
            winner = 0
            break
        if b[j] > a[j]:
            winner = 1
            break

    if winner is None:
        # No discriminating cue — pure guess.
        return np.ones(2) / 2.0

    scores = np.array([1.0, 0.0]) if winner == 0 else np.array([0.0, 1.0])

    beta = float(parameters["beta"])
    epsilon = float(parameters["epsilon"])

    # Softmax with max-subtraction for numerical stability. For the
    # binary TTB score this collapses to sigmoid(beta) for the winner,
    # giving a direct mapping from beta onto the paper's flip-noise
    # levels (beta=0 \leftrightarrow 50/50; beta \gg 1 \leftrightarrow deterministic).
    z = beta * (scores - scores.max())
    e = np.exp(z)
    p_core = e / e.sum()

    n_opts = p_core.shape[0]
    return (1.0 - epsilon) * p_core + epsilon * (np.ones(n_opts) / n_opts)
\end{Verbatim}

\textit{policy.}
\begin{Verbatim}[breaklines=true,breakanywhere=true,fontsize=\footnotesize]
def policy(probabilities):
    return int(np.argmax(probabilities))
\end{Verbatim}

\paragraph{Tallying}
\textit{Theory.} People compare two options by counting, across all features, how often one option has a higher value than the other. The option that wins on more features is chosen. Tallying discards cardinal magnitudes --- only the sign of each feature-wise comparison matters --- so the heuristic is robust to monotone rescaling of individual features and cannot be swayed by a single large feature difference in the way Equal-Weight can. Ties on an individual feature contribute nothing to either count: that cue is simply treated as uninformative for the pair. No feature is privileged, in contrast to Take The Best; every cue contributes equally to the tally. When the two counts are equal the heuristic has no basis for preference and the learner must guess. Response noise enters through a softmax over the two tallies with inverse temperature beta (interpolating between fully deterministic choice at large beta and uniform guessing at beta = 0), plus an independent lapse that with probability epsilon replaces the softmax output with a uniform pick over the two options.

\textit{Parameters.}
\begin{Verbatim}[breaklines=true,breakanywhere=true,fontsize=\footnotesize]
beta: [0.1, 20.0]
epsilon: [0.0, 0.5]
\end{Verbatim}

\textit{predict.}
\begin{Verbatim}[breaklines=true,breakanywhere=true,fontsize=\footnotesize]
def predict(parameters, stimulus, history):
    # Paper-faithful Tallying heuristic (Dawes 1979; Gigerenzer &
    # Goldstein 1999). Stimulus is the pair of option feature vectors
    # for the current trial: array-like of shape (2, n_features),
    # with row 0 = option A, row 1 = option B. History is ignored.
    stim = np.asarray(stimulus, dtype=float)
    if stim.ndim != 2 or stim.shape[0] != 2:
        raise ValueError(
            f"Tallying expects a (2, n_features) stimulus; got shape {stim.shape}."
        )

    a, b = stim[0], stim[1]
    # Count strict feature-wise wins; ties contribute to neither option.
    a_wins = float(np.sum(a > b))
    b_wins = float(np.sum(b > a))
    scores = np.array([a_wins, b_wins])

    beta = float(parameters["beta"])
    epsilon = float(parameters["epsilon"])

    # Softmax with max-subtraction for numerical stability. When
    # a_wins == b_wins the softmax is exactly uniform regardless of
    # beta, which is the correct behavior for an undiscriminating
    # tally.
    z = beta * (scores - scores.max())
    e = np.exp(z)
    p_core = e / e.sum()

    n_opts = p_core.shape[0]
    return (1.0 - epsilon) * p_core + epsilon * (np.ones(n_opts) / n_opts)
\end{Verbatim}

\textit{policy.}
\begin{Verbatim}[breaklines=true,breakanywhere=true,fontsize=\footnotesize]
def policy(probabilities):
    return int(np.argmax(probabilities))
\end{Verbatim}

\subsection*{Discovered theories}

\paragraph{Contextual Relative Advantage Normalization Theory}
\textit{Theory.} Decision-makers evaluate options by comparing feature differences, but the subjective weight of each difference is normalized by the maximum absolute difference present in the current stimulus context. Small differences are dynamically overshadowed (via a non-linear contrast effect) only when a massive advantage exists on another cue, avoiding the need for a fixed internal threshold. This preserves sensitivity to distributed small advantages when the overall context lacks extreme differences, but allows a single large advantage to dominate when present.

\textit{Parameters.}
\begin{Verbatim}[breaklines=true,breakanywhere=true,fontsize=\footnotesize]
gamma: [0.0, 1.0]
tau: [0.1, 10.0]
alpha: [0.1, 5.0]
rho: [0.0, 3.0]
beta: [0.1, 30.0]
epsilon: [0.0, 0.5]
validities: validities
\end{Verbatim}

\textit{predict.}
\begin{Verbatim}[breaklines=true,breakanywhere=true,fontsize=\footnotesize]
def predict(parameters, state, history):
    import numpy as np

    stim = np.asarray(state, dtype=float)
    val = np.asarray(parameters['validities'], dtype=float)

    gamma = float(parameters['gamma'])
    tau = float(parameters['tau'])
    alpha = float(parameters['alpha'])
    rho = float(parameters['rho'])
    beta = float(parameters['beta'])
    epsilon = float(parameters['epsilon'])

    # Normalize validities and apply shrinkage toward equal weights
    val_norm = val / np.sum(val)
    n_cues = len(val)
    w = (1.0 - gamma) * val_norm + gamma * (1.0 / n_cues)

    # Non-linear validity transformation to allow transitions toward lexicographic weighting
    w_k = w ** tau
    w_k = w_k / np.sum(w_k)

    n_options = stim.shape[0]
    scores = np.zeros(n_options)

    # Find the maximum absolute difference across all cues and all pairs in the current context
    global_max_diff = 0.0
    for i in range(n_options):
        for j in range(n_options):
            if i != j:
                global_max_diff = max(global_max_diff, np.max(np.abs(stim[i] - stim[j])))

    for i in range(n_options):
        for j in range(n_options):
            if i != j:
                diff = stim[i] - stim[j]
                if global_max_diff > 0:
                    # Normalize differences by the maximum contextual difference
                    norm_diff = np.abs(diff) / global_max_diff

                    # Non-linear contrast effect: alpha > 1 suppresses smaller relative differences
                    salience = norm_diff ** alpha

                    # Scale back by global_max_diff ** rho to preserve absolute magnitude sensitivity
                    evidence = w_k * np.sign(diff) * salience * (global_max_diff ** rho)
                    scores[i] += np.sum(evidence)

    # Softmax over the scores
    z = beta * scores
    z -= np.max(z)  # for numerical stability
    e = np.exp(z)
    p = e / np.sum(e)

    # Incorporate lapse rate
    return (1.0 - epsilon) * p + epsilon * (np.ones(n_options) / n_options)
\end{Verbatim}

\textit{policy.}
\begin{Verbatim}[breaklines=true,breakanywhere=true,fontsize=\footnotesize]
def policy(probs):
    import numpy as np
    probs = np.asarray(probs, dtype=np.float64)
    probs /= probs.sum()
    return int(np.random.choice(len(probs), p=probs))
\end{Verbatim}

\paragraph{Relative Contextual Salience Lexicographic Theory}

\textit{Theory.} For each pair of options, cues are evaluated in order of their validity. Instead of making a deterministic stop/go decision, each cue has a probability of terminating the search based on how strongly it discriminates between the options. Evidence is accumulated as the expected contribution across all possible stopping points. If no cue triggers a stop, the remaining evidence is integrated using a compensatory weighted combination of all cues. The resulting evidence is accumulated into an overall score for each option, which is converted into choice probabilities using a softmax function.

\textit{Parameters.}
\begin{Verbatim}[breaklines=true,breakanywhere=true,fontsize=\footnotesize]
alpha: [0.1, 10.0]
theta: [0.0, 5.0]
beta: [0.1, 20.0]
gamma: [0.0, 1.0]
epsilon: [0.0, 0.5]
lambda_stop: [0.0, 1.0]
validities: validities
\end{Verbatim}

\textit{predict.}
\begin{Verbatim}[breaklines=true,breakanywhere=true,fontsize=\footnotesize]
def predict(parameters, state, history):
    import numpy as np

    stim = np.asarray(state, dtype=float)
    val = np.asarray(parameters['validities'], dtype=float)

    alpha = float(parameters['alpha'])
    theta = float(parameters['theta'])
    beta = float(parameters['beta'])
    gamma = float(parameters['gamma'])
    epsilon = float(parameters['epsilon'])
    lambda_stop = float(parameters['lambda_stop'])

    n_options, n_cues = stim.shape

    # Normalize validities and apply shrinkage
    val_norm = val / np.sum(val)
    w = (1.0 - gamma) * val_norm + gamma * (1.0 / n_cues)

    # Sort cues by validity descending
    cue_order = np.argsort(val)[::-1]

    scores = np.zeros(n_options)

    for i in range(n_options):
        for j in range(n_options):
            if i == j:
                continue

            diff = stim[i] - stim[j]
            abs_diff = np.abs(diff)
            M = np.max(abs_diff)

            if M == 0:
                continue

            r = abs_diff / (M + 1e-6)

            expected_evidence = 0.0
            p_not_stopped = 1.0

            for k in cue_order:
                if r[k] == 0:
                    h = 0.0
                else:
                    # Soft stopping probability based on relative difference to max, scaled by lambda_stop
                    h = lambda_stop * (r[k] ** alpha) / (r[k] ** alpha + theta ** alpha + 1e-9)

                p_stop = p_not_stopped * h
                expected_evidence += p_stop * np.sign(diff[k])
                p_not_stopped *= (1.0 - h)

            # Fallback to compensatory weighting incorporating relative magnitudes
            fallback_evidence = np.sum(w * diff / (M + 1e-6))
            expected_evidence += p_not_stopped * fallback_evidence

            scores[i] += expected_evidence

    # Softmax over the scores
    z = beta * scores
    z -= np.max(z)  # for numerical stability
    e = np.exp(z)
    p = e / np.sum(e)

    # Incorporate lapse rate
    return (1.0 - epsilon) * p + epsilon * (np.ones(n_options) / n_options)
\end{Verbatim}

\textit{policy.}
\begin{Verbatim}[breaklines=true,breakanywhere=true,fontsize=\footnotesize]
def policy(probs):
    import numpy as np
    probs = np.asarray(probs, dtype=np.float64)
    probs /= probs.sum()
    return int(np.random.choice(len(probs), p=probs))
\end{Verbatim}





\end{document}